\documentclass{article}

    \usepackage[final]{neurips_2024}

\usepackage{natbib}
\setcitestyle{numbers,square}
\usepackage[utf8]{inputenc} % allow utf-8 input
\usepackage[T1]{fontenc}    % use 8-bit T1 fonts
\usepackage{hyperref}       % hyperlinks
\usepackage{url}            % simple URL typesetting
\usepackage{booktabs}       % professional-quality tables
\usepackage{amsfonts}       % blackboard math symbols
\usepackage{nicefrac}       % compact symbols for 1/2, etc.
\usepackage{microtype}      % microtypography
\usepackage{xcolor}         % colors
\usepackage{graphicx} 
\usepackage{amsmath}
\usepackage{multirow}
\usepackage{amssymb}
\usepackage{caption}
\usepackage{floatrow}
\newfloatcommand{capbtabbox}{table}[][\FBwidth]
\usepackage{color}
\title{LESS: Label-Efficient and Single-Stage \\ Referring 3D Segmentation}

\author{%
  Xuexun Liu$^{1}$\thanks{Equal contributions.} , Xiaoxu Xu$^{1}$\footnotemark[1] , Jinlong Li$^{2}$\footnotemark[1] , Qiudan Zhang$^{1}$, Xu Wang$^{1}$\thanks{Corresponding author.} , Nicu Sebe$^{2}$, Lin Ma$^{3}$ \\
  $^{1}$College of Computer Science and Software Engineering, Shenzhen University, \\
  Shenzhen, 518060, China.\\
  $^{2}$University of Trento, Italy.\\
  $^{3}$Meituan Inc., China. \\
}

\begin{document}

\maketitle

\begin{abstract}
Referring 3D Segmentation is a visual-language task that segments all points of the specified object from a 3D point cloud described by a sentence of query. Previous works perform a two-stage paradigm, first conducting language-agnostic instance segmentation then matching with given text query. However, the semantic concepts from text query and visual cues are separately interacted during the training, and both instance and semantic labels for each object are required, which is time consuming and human-labor intensive. To mitigate these issues, we propose a novel Referring 3D Segmentation pipeline, \textbf{L}abel-\textbf{E}fficient and \textbf{S}ingle-\textbf{S}tage, dubbed \textbf{LESS}, which is only under the supervision of efficient binary mask. Specifically, we design a Point-Word Cross-Modal Alignment module for aligning the fine-grained features of points and textual embedding. Query Mask Predictor module and Query-Sentence Alignment module are introduced for coarse-grained alignment between masks and query. Furthermore, we propose an area regularization loss, which coarsely reduces irrelevant background predictions on a large scale. Besides, a point-to-point contrastive loss is proposed concentrating on distinguishing points with subtly similar features. Through extensive experiments, we achieve state-of-the-art performance on ScanRefer dataset by surpassing the previous methods about 3.7\% mIoU using only binary labels. Code is available at \url{https://github.com/mellody11/LESS}.

\end{abstract}

\section{Introduction}

% 第一段在应用那里找几篇论文引一引
% Given a 3D point cloud scene and a free-form natural language expression, Referring 3D Segmentation aims to identify and segment the specific object described by the expression, which allows users to interact and analyze 3D data through verbal instructions or queries. This approach is particularly beneficial in applications that necessitate direct interaction with 3D environments, such as augmented reality (AR) systems, embodied-AI, robotics, virtual reality (VR) environments, and in fields like architecture and medical imaging, where precise identification and segmentation of objects based on descriptive queries can significantly enhance user experience and operational efficiency.
% 尽管referring instance segmentation在2D图片中已经取得不错的进展，但Referring 3D Segmentation任务仍具有较大的挑战性（挑战性下一段讲）。

% Given a 3D point cloud scene and a free-form natural language expression, Referring 3D Segmentation aims to segment the specific object described by the expression. 
Referring 3D Segmentation task aims to segment the specific object from a 3D point cloud scene with a free-form natural language expression. 
It allows users to interact and analyze 3D data through verbal instructions or queries. This approach is particularly beneficial in applications that necessitate direct interaction with 3D environments, such as augmented reality (AR) systems, embodied-AI, robotics, virtual reality (VR) environments, and in fields like architecture and medical imaging, where precise identification and segmentation of objects based on descriptive queries can significantly enhance user experience and operational efficiency. 

% 这里可以参考一下lavt，看看他是怎么写的
% 论述现阶段Referring 3D Segmentation任务，包括怎么做的，用到了什么信息（围绕第一张图展开）；然后分析弊端，现阶段3D RIS利用了大量的标注数据，是time consuming and labor intensive

% Referring 3D Segmentation的主要挑战是所指向的实例物体在整个场景中的占比小，进行精确定位难度大；由于场景大背景多物体多，所以干扰也多，难以分辨目标物体。所以先前的方法采取了较为取巧的解决方案（如图一左侧所示）：采用两阶段的方法，在预分割阶段先通过一个3D Unet进行实例分割并获得每个实例的特征，这些实例会作为候选实例送入下一个阶段；在实例匹配阶段结合语言描述，将实例物体与语言描述进行匹配，相似度得分最高的物体即为最终预测物体。这样分割-匹配的策略能够很好地排除背景和无关物体的干扰，使网络更容易地筛选出正确的物体。

\begin{figure}[]
  \centering
  \includegraphics[width=\linewidth]{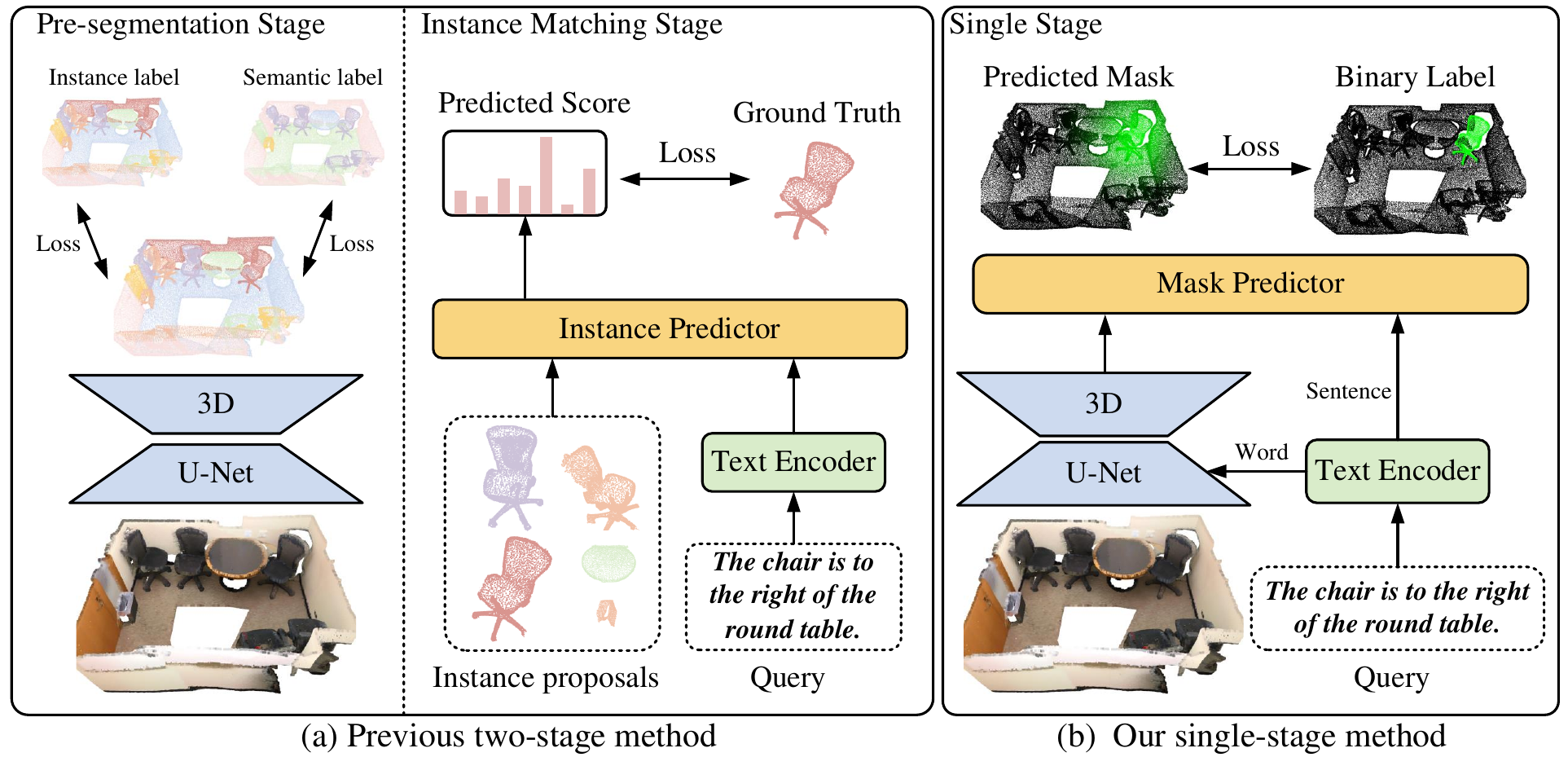}
  % 两阶段方法先进行实例分割，再结合语言描述对实例进行匹配筛选。但这样的方法需要额外的标签，且在分割阶段没有融合语言信息。
  % 单阶段方法直接利用被指向物体的点云掩码进行训练，并较早地进行语言特征融合。
  % 注意，这里为了简洁没有提到用额外标签和晚融合
  \caption{Comparison between the two-stage method and our single-stage method. (a) The two-stage method initially performs instance segmentation with instance labels then semantic labels to get the instance proposals and bases on the provided query to match the most relevant instance proposal.  (b) Our single-stage method only utilizes the binary mask of the described object for training and integrates language and vision features during feature extraction.}
  \label{fig:introfig}
\end{figure}

Previous Referring 3D Segmentation \cite{TGNN,Xrefer} mainly leverage a two-stage workaround, as shown in Fig.\ref{fig:introfig} (a). They typically adopt a 3D instance semantic segmentation network to get the instance proposals at the first stage. The predicted instance proposals will be utilized to match with the queries and finally get the final prediction mask according to the matching score. Although those method have achieved remarkable performance, there still exist some problems. First, owing to the large-scale and irregular 3D point clouds, some instance proposals may leave out the target in the pre-segmentation stage. Besides, lacking of linguistic guidance in the segmentation stage fails to focus on the objects that are more essential to the referring task. Moreover, existing Referring 3D Segmentation utilizes both instance labels and semantic labels to segment target proposal rather than binary mask used in referring image segmentation, which is more time consuming and labor intensive.

% 为了解决上面提到的问题，包括二阶段的缺陷和Referring 3D Segmentation的固有的挑战。我们提出LESS，一个标签高效的单阶段Referring 3D Segmentation方法。详细地说，为了解决multiple objects and backgrounds introduces significant interference的问题，我们提出PWCA模块进行特征在主干网络的早融合，进行点和词的细粒度对齐；采用QMA模块和QSA模块进行掩码和句的粗粒度对齐。PWCA模块中的跨模态交叉注意力将文本编码器提取到的词特征与点云特征进行对齐，随后利用强大的3D稀疏卷积层进一步提取有用的多模态信息。QMA模块利用提取的多模态特征解码可学习的查询嵌入，生成候选掩码；QSA模块将解码的查询嵌入与句特征计算相似度，并将相似度作为权重对候选掩码进行加权求和得到最终的掩码。为了解决三维点云场景庞大而被描述的物体小，难以准确定位和分割的问题，我们提出区域正则化损失函数和点-点对比损失函数。区域正则化损失通过约束预测掩码的概率值减少无关的背景预测，点-点对比损失约束正负样本在隐空间的距离实现细粒度的物体分割。

%面对multiple objects and backgrounds introduces significant interference的问题，提出PWCA模块进行特征在主干网络的早融合，进行点和词的细粒度对齐；采用QMA模块和QSA模块进行掩码和句的粗粒度对齐。面对三维点云场景庞大而被描述的物体小，难以定位和准确分割的问题，我们提出区域正则化损失函数和点-点对比损失函数。
%详细来说，PWCA模块采用跨模态交叉注意力将文本编码器提取到的词特征与点云特征进行对齐，随后利用强大的3D稀疏卷积层进一步提取有用的多模态信息；QMA模块利用提取的多模态特征解码可学习的查询嵌入，生成候选掩码；QSA模块将解码的查询嵌入与句特征计算相似度，并将相似度作为权重对候选掩码进行加权求和得到最终的掩码。

% To address the aforementioned issues, we propose \textbf{LESS}, a \textbf{L}abel \textbf{E}fficient \textbf{S}ingle \textbf{S}tage Referring 3D Segmentation method.

To address the aforementioned problems, we propose a \textbf{L}abel-\textbf{E}fficient and \textbf{S}ingle-\textbf{S}tage Referring 3D Segmentation method, namely \textbf{LESS}, which is under the supervision of binary mask, as shown in  Fig.\ref{fig:introfig} (b). We first process the query with text encoder to get the word-level feature and sentence-level feature. Then we extract the multi-modal feature with the guidance of text feature through a 3D sparse U-Net. Finally, the mask predictor aligns multi-modal features with language features, and directly predicts the mask of the described object. Here only object mask serves as the label to supervise the whole training procedure. 

However, 3D point cloud inherently provide a higher level of complexity and a large scale. There exists numerous different objects in a single 3D scene compared to the referring image segmentation task. Besides, binary mask has less semantic meanings compared to instance labels and semantic labels. These challenges make it difficult to supervise our model to localize and segment target objects with only binary mask. Therefore, we propose to alleviate these problems by some ways, as shown in Fig.\ref{fig:backbone}. Firstly, to facilitate fine-grained alignment between points and words, we propose Point-Word Cross-Modal Alignment module. The PWCA module utilizes cross-modal attention in the multi-modal context to align textual features extracted by the text encoder with point cloud features, followed by further extraction of useful multi-modal information using robust 3D sparse convolutional layers. Thus we can extract a more semantic meaning fused feature. Meanwhile, we employ Query Mask Predictor, which utilizes the extracted multi-modal features to decode learnable query embeddings, generating candidate masks. By introducing the Query-Sentence Alignment modules, we can compute similarity between the decoded query embeddings and sentence features, using the similarity as weights to perform weighted summation of the candidate masks to produce the final mask. To address the significant interference caused by multiple objects and backgrounds, we propose an area regularization loss and a point-to-point contrastive loss. Area regularization loss reduces irrelevant background predictions by constraining the probabilities of the predicted mask, while point-to-point contrastive loss constrains the distances between positive and negative points in the latent space to achieve better segmentation.

To summarize, our contributions are as follows:
\begin{itemize}
    \item We propose a new Referring 3D Segmentation method LESS, which directly performs Referring 3D Segmentation at a single stage to bridge the gap between detection and matching under the supervision of binary mask. To the best of our knowledge, LESS is the first work investigating label-efficient and single-stage in Referring 3D Segmentation task.
    \item Our LESS utilize a Point-Word Cross-Modal Alignment module to align fined-grained point and word features. Besides, we employed Query Mask Predictor module and Query-Sentence Alignment modules for coarse-grained alignment between masks and sentences. Moreover, the area regularization loss and the point-to-point contrastive loss are introduced to better support to eliminate interference caused by multiple objects and backgrounds.
    \item Extensive experiments confirm the effectiveness of our method. 
    Our method outperforms the existing state-of-the-art method on ScanRefer dataset with only the supervision of binary mask.  Our LESS and its results provide valuable insights to improve further research of label-efficient and single-stage Referring 3D Segmentation.
\end{itemize}

% \begin{itemize}
%     \item We analyze the the challenge of Referring 3D Segmentation and the issues of previous two-stage methods. We propose LESS, which is the first single-stage Referring 3D Segmentation model. 
%     \item We designed PWCA module to align fined-grained point and word features, we introduced a region regularization loss function alongside a point-to-point contrastive loss function to progressively constrain object segmentation masks from coarse to fine.
%     \item Extensive experiments confirm the effectiveness of our method. LESS outperforms the state-of-the-art method by 3.7 in terms of mIoU on \textit{ScanRefer} \cite{scanrefer} dataset without using semantic and instance labels.
% \end{itemize}

\section{Related Works}

\subsection{Referring 3D Segmentation}

Referring 3D Segmentation has previously received limited exploration, however, with the advancements in multi-modal learning and embodied AI, it is set to attract increasing interest in the future. TGNN \cite{TGNN} is the first to introduce Referring 3D Segmentation task. They initially trained an instance segmentation network, followed by a Graph Neural Network (GNN) to learn features of instances and their relationships guided by linguistic information. Building on TGNN, X-RefSeg \cite{Xrefer} developed a cross-modal graph. They employed an GNN to model the texture and spatial relationships of instances, and refining the results through inference and matching processes.

\subsection{3D Visual Grounding}

3D visual grounding aims to locate the object within point clouds mentioned by free-form natural language descriptions. Most methods follow a two-stage detection-then-matching pipeline. Initially, they utilize a pre-trained 3D detector \cite{votenet, groupfree} or segmenter \cite{pointgroup,softgroup} to extract object representations. Subsequently, these methods align text features with object features to identify the best-matched object. Researchers primarily concentrate on the second stage which involves modeling object relationships and exploring feature fusion between language and objects. Methods employed include multi-modal feature concatenation \cite{scanrefer,referit3d}, attention-based multi-modal feature alignment \cite{3DVGtransformer, transrefer3d}, graph neural network-based reasoning \cite{TGNN,instancerefer}, and the alignment of visual and language features aided by 2D images \cite{sat, multi3drefer, weakly, xu20243d}.Other researchers have investigated single-stage approaches for 3D visual grounding. 3D-SPS \cite{3dsps} views the task as key point selection, progressively identifying keypoints with the guidance of language and directly locates the target. BUTD-DETR \cite{bottomup} employs a transformer decoder \cite{detr} to identify described objects using language cues and proposal boxes. Building on this, EDA \cite{eda} enhances dense alignment between objects and point clouds by explicitly decoupling textual attributes from sentences.

The distinction between 3D visual grounding and Referring 3D Segmentation lies in the latter's enhanced localization precision, offering significant value in applications like robotic grasping. While traditional single-stage 3D visual grounding methods regress a 3D bounding box, our single-stage approach decodes a binary mask for the entire point cloud scene, presenting a more complex challenge.

\subsection{Referring Image Segmentation}

Referring image segmentation is a visual task that involves pixel-level segmentation of an image's target object based on a referring expression. Early approaches \cite{2dsegmentationfrom,2dreferring,li2022weakly,2dsee,li2022expansion} utilize CNNs and RNNs to extract features and then perform simple concatenation for multi-modal feature fusion. Subsequent research \cite{2drestr,2dreferringtransformer,2dvlt} adopt transformer models for more effective feature extraction and fusion. Recent studies \cite{lavt,2dencoder,coupalign} focus on identifying optimal positions for language-vision alignment. Additionally, some methods \cite{cris, maskgroup} have enhanced alignment between language and pixel features by designing specialized loss functions. Other approaches \cite{2dpoly,2dseqtr} treat referring image segmentation as an auto-regressive vector generation problem, creating masks from generated closed vectors. 

Methods for referring image segmentation have been extensively explored, yet they cannot be directly applied to Referring 3D Segmentation due to the inherent challenges of point cloud scenes. Additionally, unlike referring image segmentation methods that typically employ pre-trained visual encoders, Referring 3D Segmentation lacks such resources, compelling reliance on limited supervisory signals for model training.

\section{Method}

\begin{figure}[]
  \centering
  \includegraphics[width=\linewidth]{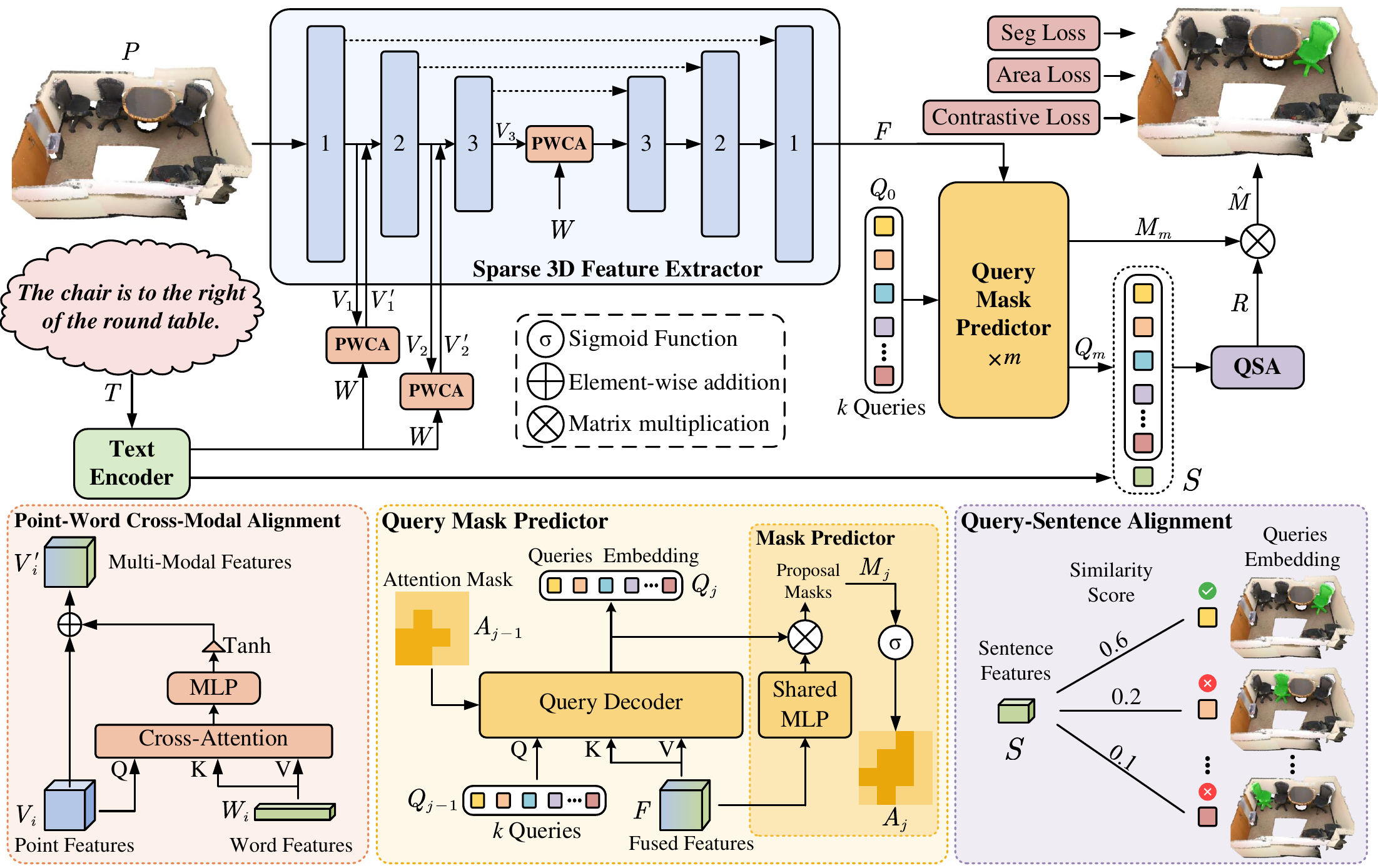}
  % \caption{Overview of our \textbf{LESS} framework. Given a point cloud scene $P$
  % and a query $T$, we use a text encoder and a sparse 3D feature extractor to extract word features $F_{w}$ and sentence features $F_{s}$. Meanwhile, the PWCA module aligns the word features $F_{w}$ with multi-scale point cloud features $\{F_{i}\}_{i=1}^{3}$ extracted by a sparse 3D feature extractor. After that, we use an m-layer QMP module to decode $K$ learnable queries $Q_{0}$ and fused feature $F$ to query embeddings $Q_{m}$ and proposal masks $M_{m}$. Finally, sentence features $F_{s}$ and query embeddings $Q_{m}$ are utilized to compute the similarity scores $S$ in QSA module and filter the proposal masks $M_{m}$ to the final mask prediction $\hat{M}$.}
  \caption{Overview of our \textbf{LESS} framework. Given a point cloud scene $P$, we use a sparse 3D feature extractor to extract multi-scale feature $V_{i}$. The query $T$ is sent to a text encoder and we obtain the word features $W$ and sentence features $S$. Meanwhile, we introduce a PWCA module aligns the word features $W$ with the multi-scale point cloud features $V_{i}$. After that, an $m$-layer QMP module is adopted to decode $K$ learnable queries $Q_{0}$ base on the fused feature $F$, and output query embeddings $Q_{m}$ and proposal masks $M_{m}$. Finally, QSA module aligns the query embeddings $Q_{m}$ with sentence features $S$, i.e., computes the similarity scores $R$ that filter the proposal masks $M_{m}$ to the final mask prediction $\hat{M}$.}
  \label{fig:backbone}
\end{figure}

The overall framework of our proposed LESS is shown in Fig.\ref{fig:backbone}, which leverages Point-Word Cross-Modal Alignment and Query-Sentence Alignment to facilitate multi-modal interaction. In this section, we start by introducing our visual and text feature extractor in Sec.\ref{section:Encoder}. Then Query Mask Predictor and Query-Sentence Alignment is detailed in Sec.\ref{section:QMP} and Sec.\ref{section:QSA}. Finally in Sec.\ref{section:Loss}, we introduce our area regularization loss and point-to-point contrastive loss.

% As depicted in Fig.\ref{fig:backbone}, given a descriptive query sentence $T \in \mathbb{R}^{L}$ and an indoor point cloud scene $P \in \mathbb{R}^{N \times 6}$—where $L$ represents the sentence length and $N$ denotes the number of points, each with $xyz$ and $rgb$ attributes—Referring 3D Segmentation aims to predict a binary mask $\hat{Y} \in \{0,1\}^{N}$ corresponding to the object described by the sentence. In the training phase, prior works \cite{TGNN,Xrefer} initially train an instance segmentation network supervised by instance labels $Y_{ins} \in \mathbb{R}^{N}$ and semantic labels $Y_{sem} \in \mathbb{R}^{N}$ for each scene. In contrast, our method solely relies on a binary mask label $Y \in \{0,1\}^{N}$ for training, eliminating the need for instance or semantic labels.

% We employ a sparse 3D feature extractor and a text encoder to extract the features of $P$ and $T$, and perform multi-modal alignment through the point-word cross-modal alignment (PWCA) module as described in Sec.\ref{section:Encoder}. For proposal mask generation, a query mask predictor with $N$ learnable queries is utilized in Sec.\ref{section:QMP}. The query embeddings align with sentence features in Sec.\ref{section:QSA} to obtain the final mask prediction through matrix multiplication. Additionally, we introduce our area regularization and point-to-point contrastive learning in Sec.\ref{section:Loss}.

% 目前的架构图

% ---------------------------------------------------------------模块开始--------------------------------------------------------------

\subsection{Visual and Text Feature Extractor}
\label{section:Encoder}

\paragraph{Sparse 3D Feature Extractor.} The point cloud scene $P \in \mathbb{R}^{N \times 6}$ contains $N$ points in the scene, and each point is represented with six dimensions of RGBXYZ attributes. We first voxelize points into regular voxels and adopt a sparse 3D U-Net \cite{spformer, pointgroup} to extract point-wise fused feature $F \in \mathbb{R}^{N\times C}$. Here the encoder part of the U-Net has $5$ stages and the feature from the $i$-th encoder stage is denoted as $V_{i}$.
% we leverage all these multi-scale features, denoted as $\{F_{i}\in \mathbb{R}^{N_i \times C_i}\}_{i=1}^{5}$, for further feature fusion.

\paragraph{Text Encoder.}
Given the query sentence $T$ with $L$ words, a text encoder is used to  embed the query into $C$-dimensional feature vectors. In this paper we choose GRU \cite{gru}, BETR \cite{bert} and RoBERTa \cite{roberta} as our text encoder respectively and fine-tune the BERT or RoBERTa during training. Finally, we can get both word features $W \in \mathbb{R}^{L\times C}$  and sentence features $S \in \mathbb{R}^{C}$ after the text encoder.

% Given a specific language description, we first utilize a pre-trained tokenizer \cite{tokenizer} to tokenize the word into a word embedding representation. we adopt BERT\cite{bert} as text encoder and encode the word embedding into word features $F_{w} \in \mathbb{R}^{L\times C}$  and sentence features $F_{s}\in \mathbb{R}^{C}$. 

\paragraph{Point-Word Cross-Modal Alignment.} Due to the lack of such rich annotations as \cite{TGNN, Xrefer}, it is crucial for our model to learn the relationship between fine-grained word-level features and point-level features in such a point-level segmentation task. Meanwhile, we notice that leveraging the rich convolutional layers in the encoder to excavate multi-modal context is effective way to extract language-aware visual feature. Therefore, we design Point-Word Cross-Modal Alignment (PWCA) module, as shown in the lower left part in Fig.\ref{fig:backbone}, which contains a standard cross-attention module \cite{transformer} and an MLP-Tanh gate. Cross-attention module aligns point-wise and word-wise feature to get the language-aware visual feature. A nonlinear tanh gate is adopted to prevent the fused signal from overwhelming the original signal. Given the multi-scale point cloud features $\{V_{i} \in \mathbb{R}^{N_{i}\times C_{i}}\}_{i=1}^{5}$, we simply project the word feature  $W \in \mathbb{R}^{L \times C}$ to $W_{i} \in \mathbb{R}^{L\times C_{i}}$. PWCA can be formulated as follow:
\begin{align}
    V_{i}^{\prime} = \text{Tanh}(\text{MLP}(\text{CrossAttn}(V_{i}, W_{i})) + V_{i}, \quad   i\in \{1,...,5\},
\end{align}
where $i$ indicates the $i$-th stage of the encoder part of our sparse 3D feature extractor. Here we use $V_{i}$ as the query and $W_{i}$ as the key and value for cross attention.

\subsection{Query Mask Predictor}
\label{section:QMP}
Inspired by \cite{spformer, mask3d}, 
Query Mask Predictor (QMP) module, as shown in Fig.\ref{fig:backbone}, takes fused feature $F \in \mathbb{R}^{N \times C}$ and learnable queries $Q_{0} \in\mathbb{R}^{K \times C}$ as input and progressively distinguishes the referring target by multi-layer cross-modal transformers. Finally, we extract the proposal masks $M_{m} \in \mathbb{R}^{K\times N}$ based on queries embeddings $Q_{m}\in\mathbb{R}^{K \times C}$ and fused feature $F$.

% % 开头加一些功能描述性的语句
% As shown in Fig.\ref{fig:backbone}, a multi-layer stacked query mask predictor is designed to iteratively produce query embedding $Q_{m}\in\mathbb{R}^{K \times D}$ and proposal mask prediction $M_{m} \in \mathbb{R}^{K\times N}$ by using a query decoder  and a mask predictor. Where $K$ denotes the number of queries, $D$ denotes feature dimensions in the query decoder, and $m$ denotes the number of layers, all queries in each layer has the same dimension $\mathbb{R}^{K \times D}$. 

\paragraph{Query Decoder.} 

% Taking the $j$-th ( $j\in \{1,2,3...,m\}$ ) layer as an example . With the fused feature $F$ from sparse 3D feature extractor in Sec.\ref{section:1}, $K$ learnable queries $Q_{j-1}$ from former layer capture multi-modal contextual information via a query decoder which contains a masked cross-attention layer, self-attention layer and a feed forward layer as in  \cite{transformer}.  The attention mask $A_{j-1} \in \{0, 1\}^{K\times N}$ is calculated by the former layer and the query embedding $Q_{j}$ is sent to next layer for further learning.
% We randomly initialize the query $Q_{0}$ at the first layer and exchange the order of the cross-attention and self-attention layer. It is because the learnable query $Q_{0}$ does not initially carry multi-modal feature information and requires early interaction with multi-modal features $F$ to facilitate a more effective acquisition of multi-modal knowledge.

As illustrated in the lower middle section of Fig.\ref{fig:backbone}. Here query decoder is comprised $m$-layer masked cross-attention modules \cite{transformer}, where the fused features $F$ are served as keys and values and $A_{j-1} \in \{0, 1\}^{K \times N}$ is utilized as the attention mask. Therefore, in the $j$-th layer, the learnable queries $Q_{j-1}$ capture multi-modal contextual information from fused feature $F$ via a query decoder to obtain the query embeddings $Q_{j} \in \mathbb{R}^{K \times C}$.

% As illustrated in the lower middle section of Fig.\ref{fig:backbone}, in the $j$-th layer ($j \in \{1, 2, 3, ..., m\}$), $K$ learnable queries $Q_{j-1}$ capture multi-modal contextual information $F$ via a Query Decoder to obtain the query embeddings $Q_{j}$. The decoder comprises a masked cross-attention layer, a self-attention layer, and a feed-forward layer, as described in \cite{transformer}. The attention mask $A_{j-1} \in \{0, 1\}^{K \times N}$ is computed by the mask predictor from the previous layer, and the query embeddings $Q_{j}$ are forwarded to the subsequent layer for further learning.

% 与spformer的区别：spformer的perdiction head需要预测语义lebel、分数、mask，我们这里只预测mask，不需要分数，所以mask只是一些proposal mask，到后面QSA模块才根据相似度确定最终的那个mask
\paragraph{Mask Predictor.}  

% After the $j$-th query decoder layer, we have query embedding $Q_{j}$, we project the fused feature $F \in \mathbb{R}^{N\times C}$ to $F^{\prime}\in \mathbb{R}^{N\times D}$ via a shared-MLP and multiplied by $Q_j \in \mathbb{R}^{K\times D}$ to generate $K$ proposal mask prediction $M_{j}\in \mathbb{R}^{K\times N}$. Finally, after a $\text{sigmoid}$ function and a threshold of $0.5$, the binary attention mask $A_{j}$ is generated. Note that, we use $Q_{0}$ and $F$ to calculate $A_{0}$ before the first layer.

First the fused feature $F \in \mathbb{R}^{N \times C}$ is processed by a Shared MLP, which indicates each Query Mask Predictor layer shares the same MLP. We perform the matrix multiplication on the new $F$ and $Q_{j} \in \mathbb{R}^{K \times C}$ to generate proposal mask predictions $M_{j} \in \mathbb{R}^{K \times N}$. After applying a sigmoid function whose threshold of 0.5, we can get the new binary attention mask $A_{j} \in \{0, 1\}^{K \times N}$.

Finally, the Query Mask Predictor is formally described as follows:
\begin{align}
    Q_{j} &= \text{Query Decoder}_{j}(Q_{j-1}, A_{j-1}, F),\\
    M_{j} &= Q_{j} \otimes \text{Shared-MLP}(F)^\top, \label{func:3} \\  
    A_{j} &= 
    \begin{cases} 
    1& \sigma(M_{j}) \geq 0.5 \\
    0& \text{otherwise}         \label{func:4}
    \end{cases},
\end{align}
where $\top$ denotes the transpose operation, $\otimes$ represents matrix multiplication, and $\sigma$ refers to the sigmoid function. Here we need to note that the $Q_{0}$ is randomly initialized. Therefore we can leverage the function (\ref{func:3}-\ref{func:4}) to initialize the attention mask $A_{0}$.

\subsection{Query-Sentence Alignment}
\label{section:QSA}
The Referring 3D Segmentation task involves the segmentation of a solitary target object according to the query. Previous modules focus on extracting language-aware visual feature from aligning point-wise feature and word-wise feature, which lacks a comprehensive perception of the entire query sentence. Therefore, we adopt Query-Sentence Alignment (QSA) to better align the query feature with sentence-level feature. We perform the  matrix multiplication on $Q_{m} \in \mathbb{R}^{K \times C}$ and $S \in \mathbb{R}^{C}$ to get the their similarity score $R \in [0,1]^{K}$. Formally,  Query-Sentence Alignment can be  represented as:
% Although the query embedding $Q_m$ is already aligned with multi-modal information $F$, incorporating both point cloud features and word features, it still lacks a comprehensive perception of the entire sentence. Therefore, we adopt Query-Sentence Alignment (QSA) to better align the query embedding with sentence-level features. We first project the sentence feature $F_{s} \in \mathbb{R}^{C}$  to $F_{s}^{\prime} \in \mathbb{R}^{D}$ via an MLP, and then align $Q_{m} \in \mathbb{R}^{K \times D}$ and $F_{s}^{\prime} \in \mathbb{R}^{D}$ by calculating their similarity score $S \in [0,1]^{K}$. Different from \cite{coupalign},   we use the feature map of the same scale as  $P$  to produce proposed mask prediction $M_m$ to provide fine-grained segmentation boundaries and reduce errors caused by upsampling operation. Formally,  Query-Sentence alignment can be  represented as:
\begin{align}
    R = \text{Softmax}&(Q_{m} \otimes S).
\end{align}
The final mask prediction $\hat{M} \in \mathbb{R}^{N}$ is produced by weighted sum of similarity score $R$ and proposed mask prediction $M_{m} \in \mathbb{R}^{K\times N}$. Finally, we use a sigmoid function and a threshold of $0.5$ to produce the final predicted binary mask $\hat{Y} \in \{0,1\}^{N}$ : 
\begin{align}
    \hat{M} = R \otimes M_{m}, \quad\quad\quad\hat{Y} =    
    \begin{cases} 
    1& \sigma(\hat{M}) \geq 0.5 \\
    0& \text{otherwise}
    \end{cases}.
\end{align}

\subsection{Loss Function}
\label{section:Loss}

\paragraph{Segmentation Loss.}
Different from previous work \cite{TGNN, Xrefer}, we take the Referring 3D Segmentation task as segmentation task with only binary mask $Y \in \{0,1\}^{N}$. Here we utilize the Binary Cross-Entropy (BCE) loss function to compute the segmentation loss, which can be formulated as:
% 与前面的工作将refering 3D instance seg建模为实例级分类任务不同，我们将refering 3D instance seg 任务建模为点级的分类任务，即将点分类为前景和背景。所以我们采用BCE损失函数计算分类损失，即
% Different from previous work \cite{TGNN, Xrefer} that models the Referring 3D Segmentation as an instance-level classification task, we treat it as a point-level classification task, which classifying points as foreground or background. Therefore, we utilize the Binary Cross-Entropy (BCE) loss function to compute the segmentation loss, which can be formulated as:
\begin{align}
 \mathcal{L}_{\text{seg}} = \text{BCE}(\sigma(\hat{M}), Y).
\end{align}

\paragraph{Area Regularization Loss.} 
% 这个故事看看怎么讲
% 打算可视化消融的一些样例，看看这个loss单独作用的效果是什么，然后结合效果讲讲故事，效果图也可以放在后面消融的部分。

%  在Referring 3D Segmentation任务中，由于一句描述仅对应点云场景中的一个物体，被描述的物体的尺寸占整个点云场景的尺寸小，所以对于大部分的物体，网络往往只能进行大致的定位，分割结果往往包含大量背景区域。为了减少背景区域面积，我们希望网络对于最终预测的掩码尺寸尽可能地小，通过与分割loss的配合，只分割最有可能的区域，而少分割无关的背景。所以，我们提出区域正则化损失，该损失旨在最小化每一个点的输出概率
For Referring 3D Segmentation task, each query always corresponds to one target object in the point cloud scene. The target objects occupy a smaller area in the large scale of 3D point cloud. As a result, the predicted mask often includes backgrounds or other objects. To address this, we propose a region regularization loss, which promotes the network to predict a smallest mask by minimize the output probability of each point, formulated as:
% For Referring 3D Segmentation task, each querycorresponds to only one object in the point cloud scene, and the scale of the described object is much smaller than that of the entire point cloud. As a result, for most objects, the network can only perform rough masks, which includes backgrounds or other objects. To address this, we propose a region regularization loss, which promotes the network to predict a smallest mask, by minimize the output probability of each point, formulated as:
\begin{align}
    \mathcal{L_{\text{area}}} = \frac{1}{N} \sum_{i=1}^{N} \sigma(\hat{M_{i}}).
\end{align}
By combining with the segmentation loss, we intend to segment only the most probable regions while reducing segmentation of large-scale irrelevant background areas. 

\paragraph{Point-to-Point Contrastive Loss.}
%  区域正则化损失一视同仁地惩罚所有点的预测概率，这可以减少大部分的背景，但对于被指向物体的邻近的背景点，由于其在特征空间中与被指物体的点非常相近，网络难以对其进行分类，因此我们提出点-点对比损失函数，在特征空间中拉近正样本点与正样本点之间的距离，推远正样本点与负样本点之间的距离，使得网络更好地分辨被指代的物体和邻近的背景点。具体而言
% 公式

Area regularization loss uniformly penalizes the predicted probabilities of all points, which can reduce the majority of the background points. However, the network struggles to differentiate between objects that possess characteristics similar to those described target object in the latent space. Therefore, we propose a point-to-point contrastive loss \cite{infonce} that pull the points from the described object together and push away the rest points: 
\begin{align}
 \mathcal{L}_{\text{p2p}} = - \frac{1}{|\mathcal{P}|} \sum_{i=1}^{|\mathcal{P}|} &\frac{\exp(\mathbf{P}_{i} \cdot \mathbf{P}_{\text{avg}} / \tau)}{\exp(\mathbf{P}_{i} \cdot \mathbf{P}_{\text{avg}} / \tau) + \sum_{j=1}^{|\mathcal{N}|} \exp(\mathbf{P}_{i} \cdot \mathbf{N}_{j} / \tau)},
\end{align}
where  $\mathcal{P}$ is the positive point set from the described object, $\mathcal{N}$ is the negative point set from the background, $\mathbf{P}_{i}$  denotes the L2-normalized feature vector of $i$ -th positive points from $F$, while $\mathbf{N}_{j}$ denotes the L2-normalized feature vector of $j$-th negative points from $F$, $\mathbf{P}_{\text{avg}}$ is the average feature vector of positive point $\mathbf{P}_{\text{avg}}=\frac{1}{\mathcal{P}} \sum_{i=1}^{|\mathcal{P}|} \mathbf{P}_{i}$, and $\tau$ is the hyper-parameter.  The contrastive loss promotes the network to distinguish the described object from the adjacent background points in a fined-grained manner. 

The overall loss function is the weighted sum of the aforementioned three loss functions:
\begin{align}
 \mathcal{L} = \lambda_{\text{seg}}\mathcal{L}_{\text{seg}} + \lambda_{\text{area}}\mathcal{L}_{\text{area}} + \lambda_{\text{p2p}}\mathcal{L}_{\text{p2p}},
\end{align}
where $\{ \lambda_{\text{seg}},\lambda_{\text{area}},\lambda_{\text{p2p}}\}$ is set to $\{ 1,1,0.05\}$ in practice to balance the contrastive loss because of the large amount of points.

\section{Experiments}

\subsection{Dataset and Experiment Settings}

\paragraph{ScanRefer.} ScanRefer \cite{scanrefer} is a dataset for 3D referring expression comprehension tasks such as 3D visual grounding and 3D referring instance segmentation. It contains $51, 583$ queries of $11, 046$ objects from $800$ ScanNet \cite{scannet} scenes. Each scene contains $13.81$ objects and $64.48$ queries on average.
% It is divided into two distinct subsets: "\textit{Unique}" and "\textit{Multiple}" which indicate that whether the scene contains more than two distractors. Previous work \cite{TGNN,Xrefer} have not reported performance on these two subsets, we report our result in the appendix.

\paragraph{Evaluation Metric.} Following previous work \cite{TGNN, Xrefer}, we use the mean intersection-over-union (\textbf{mIoU}), and \textbf{Acc@kIOU} as the evaluation metrics. The mIoU is the average of the IoU over all test samples and the Acc@kIOU measures the accuracy of test samples with an IoU higher than the threshold $k$, where $k\in \{0.25, 0.5\}$. We use A@25 and A@50 for brevity in some of the following tables.

\paragraph{Implementation Details.} We adopt the 3D spares U-Net \cite{spformer} as our 3D feature extractor. we explore multiple text encoders, i.e., GRU \cite{gru}, BERT \cite{bert} and RoBERTa \cite{roberta} in our experiments for  comparative analysis. For BERT and RoBERTa, we use the official pre-trained weights and fine-tune them during training. We set the number of queries $K$ to 20 and use a single layer for QMP module. We set an initial learning rate of 2e-5 for the text encoder and 1e-4 for the others. We reduce the learning rate by a multiplicative factor of 0.95 each epoch and adopt Adam \cite{adamw} as our optimizer. The weights of our loss function $\{ \lambda_{\text{seg}}, \lambda_{\text{area}}, \lambda_{\text{p2p}}\}$ is set to $\{ 1,1,0.05\}$. We train for 64 epochs with a batchsize of 14, and all experiments are implemented on PyTorch \cite{pytorch}.
 
\subsection{Comparison with State-of-the-Art Methods}

\begin{table}[!t]
\caption{Quantitative results of different methods on ScanRefer \cite{scanrefer} validation set. ``Supervision'' indicates the type of supervision. \textbf{Ins.} denotes instance labels and \textbf{Sem.} indicates semantic labels. \textbf{Mask} represents binary labels. \textbf{Bold} indicates the best.}
\label{tab:main_exp}
\centering
\scalebox{0.95}{
\begin{tabular}{ccclcccc}
\toprule
\multicolumn{1}{l}{} & {\color[HTML]{333333} Method} & {\color[HTML]{333333} Backbone} & Label Effort$\ddagger$ & {\color[HTML]{333333} Supervision} & {\color[HTML]{333333} mIoU} & {\color[HTML]{333333} Acc@0.25} & {\color[HTML]{333333} Acc@0.5} \\ \midrule
 & {\color[HTML]{333333} TGNN} & {\color[HTML]{333333} GRU} & & {\color[HTML]{333333} Ins.+ Sem.} & 26.10 & 35.00 & 29.00 \\
 & {\color[HTML]{333333} TGNN} & {\color[HTML]{333333} BERT} & & {\color[HTML]{333333} Ins.+ Sem.} & 27.80 & 37.50 & 31.40 \\
 & {\color[HTML]{333333} X-RefSeg} & {\color[HTML]{333333} GRU} & & {\color[HTML]{333333} Ins.+ Sem.} & 29.77 & 39.85 & 33.52 \\
\multirow{-4}{*}{\begin{tabular}[c]{@{}c@{}}Two\\ Stage\end{tabular}} & {\color[HTML]{333333} X-RefSeg} & {\color[HTML]{333333} BERT} & \multirow{-4}{*}{$>20$ min} & {\color[HTML]{333333} Ins.+ Sem.} & 29.94 & 40.33 & 33.77 \\ \midrule
 & {\color[HTML]{333333} LESS (ours)} & {\color[HTML]{333333} GRU} & & {\color[HTML]{333333} Mask} & 32.19 & 51.00 & 26.41 \\
 & {\color[HTML]{333333} LESS (ours)} & {\color[HTML]{333333} BERT} & & {\color[HTML]{333333} Mask} & 32.44 & 51.41 & 29.02 \\
\multirow{-3}{*}{\begin{tabular}[c]{@{}c@{}}Single\\ Stage\end{tabular}} & {\color[HTML]{333333} LESS (ours)} & {\color[HTML]{333333} RoBERTa} & \multirow{-3}{*}{$<2$ min} & {\color[HTML]{333333} Mask} & {\color[HTML]{333333} \textbf{33.74}} & {\color[HTML]{333333} \textbf{53.23}} & {\color[HTML]{333333} \textbf{29.88}} \\ \bottomrule
\multicolumn{8}{l}{\small $\ddagger$ The evaluate of label effort is base on a single sample.}\\
\end{tabular}}
\end{table}

% 我们将本文提出的一阶段方法与先前的二阶段方法进行比较。如表1所示，在使用相同的语言主干下，我们的方法在mIoU和Acc@0.25上分别优于先前最好的方法xxx,xxx(gru)和xxx,xxx(bert)。我们还使用了额外的语言backbone进行实验，在mIoU和Acc@0.25分别优于先前最好的方法3.8%,12.9%。值得注意的是，先前的二阶段方法需要先训练一个实例分割网络，需要实例标签和语义标签进行监督；而我们的一阶段方法直接用被描述物体的二值掩码进行训练。

% 值得注意的是，由于先前的方法采用segmentation-matching的策略，他们是对单个完整的实例进行匹配，这使得他们的方法在Acc@50的指标上有额外的优势。而我们的单阶段方法是类别、实例不可知的，当网络对物体的位置、数目Ambiguous时，会同时预测多个区域，造成准确率的下降。

As shown in Tab.\ref{tab:main_exp}, we evaluate our LESS against the previous two-stage methods. LESS outperforms the previous SOTA method using GRU \cite{gru} and BERT \cite{bert} by an impressive progress of 2.42\%, and 2.50\% on mIoU, and 11.15\% and 11.08\% on Acc@0.25 respectively. Moreover, we conduct an extra experiment using RoBERTa \cite{roberta} , outperforming the best method of 3.8\% and 12.9\% on mIoU and Acc@0.25 respectively. Such results demonstrate the potential of our method. We also reported the comparison between our method and the two-stage approach in terms of label effort. We find that our label-efficient method saves more than 90\% label effort compared to existing methods.

However we find that the performance of our LESS has a gap on Acc@0.5 compared to previous methods. Previous methods employed a segmentation-matching strategy. Once matching successfully, the IoU between predicted mask and ground truth is mostly higher than 0.5, which is beneficial for the Acc@0.5. In contrast, our single-stage method without instance labels and semantic labels and can not extract the more accurate instance candidates as prior knowledge to assist R3DIS task. Therefore, it is acceptable for our method to perform lower than previous methods on Acc@0.5 with fewer supervisory signals.

% In addition, our network shows robustness across different language backbones, with performance gains when switching to a more effective language model. 
% Note that previous methods employed a segmentation-matching strategy, which conferred an additional benefit in the Acc@50 metric. In contrast, our single-stage method is class and instance agnostic.

% which leads to occasionally predict multiple regions and cause a decrease in Acc@50 when the network is ambiguous about the position and number of objects.

% 模型的可视化可能要放附加材料了，因为来不及可视化前两个工作，所以可视化可能就没有比较的流程了，就单纯的可视化
% 到时候选一些iou比较高的multiple样例
% \subsection{Visualizations}

% % 可视化一些成功和失败的样例，与XRefSeg作对比 2*6的一个图（第一列gt，第二列Xrefseg，第三列ours，四到六列重复，第二行重复）

% In Fig.3, we.....

\subsection{Ablation Studies.}

% -------------------------------------------------双栏表格开始------------------------------

\begin{table*}
\begin{floatrow}
\capbtabbox{
\scalebox{0.90}{
\begin{tabular}{ccc|ccc}
\toprule
& PWCA & QSA & mIoU & A@25 & A@50 \\ \midrule
(a)& & & 32.66 & 51.71 & 27.20 \\
(b)& \checkmark &  & 33.44 & 52.73 & 28.92 \\
(c)& \checkmark & \checkmark & \textbf{33.74} & \textbf{53.23} & \textbf{29.88} \\ 
\bottomrule
\end{tabular}
}}{
 \caption{Module ablation on ScanRefer dataset.}
 \label{tab:module_abl}
}
\capbtabbox{
\scalebox{0.85}{
% \begin{tabular}{cc|ccc}
% \toprule
%  $\mathcal{L}_{a.}$ & $\mathcal{L}_{p.}$ & mIoU & Acc@25 & Acc@50 \\ \midrule
%  & & 25.86 & 40.85 & 16.81 \\
% \checkmark & & 31.04 \textcolor[rgb]{0.05, 0.54, 0.2588}{(+5.18)} & 49.61 \textcolor[rgb]{0.05, 0.54, 0.2588}{(+8.76)} & 24.72 \textcolor[rgb]{0.05, 0.54, 0.2588}{(+7.91)} \\
%  \checkmark & \checkmark & \textbf{33.74 \textcolor[rgb]{0.05, 0.54, 0.2588}{(+2.70)}} & \textbf{53.23 \textcolor[rgb]{0.05, 0.54, 0.2588}{(+3.62)}} & \textbf{29.88 \textcolor[rgb]{0.05, 0.54, 0.2588}{(+5.16)}} \\ 
% \bottomrule
% \end{tabular}

\begin{tabular}{ccc|ccc}
\toprule
& $\mathcal{L}_{area}$ & $\mathcal{L}_{p2p}$ & mIoU & A@25 & A@50 \\ \midrule
(a)& & & 25.86 & 40.85 & 16.81 \\
(b)&\checkmark & & 31.04 & 49.61  & 24.72  \\
(c)&\checkmark & \checkmark & \textbf{33.74} & \textbf{53.23} & \textbf{29.88} \\ 
\bottomrule
\end{tabular}

}}{
 \caption{Loss ablation on ScanRefer dataset. }
 \label{tab:loss_abl}
 \small
}
\end{floatrow}
\end{table*}

% -------------------------------------------------双栏表格结束------------------------------

\begin{figure}[!t]
  \centering
  \includegraphics[width=\linewidth]{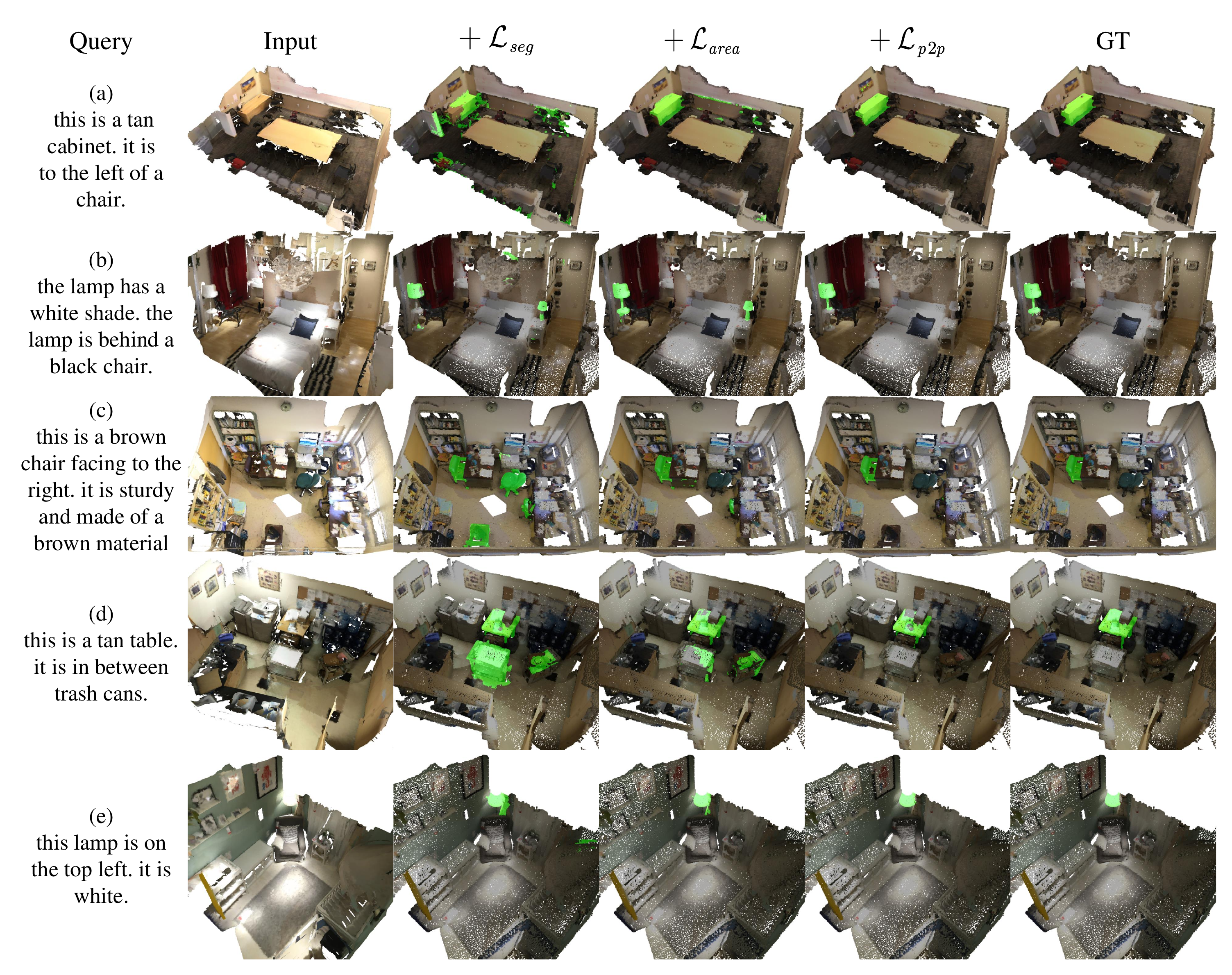}
  \caption{Final predictions using different combinations of loss functions. The queries and input scenes are shown in column 1 and 2. Columns 3 to 5 indicate the gradual addition of loss functions.}
  \label{fig:ablation}
\end{figure}

We conducted several ablation studies to evaluate the effectiveness of the key components in our proposed network. In these studies, all other components and hyper-parameter settings are kept consistent with the aforementioned experiments, except for the component being ablated.

\paragraph{Module Ablation.} 

% we regard the network containing only QMP module as the baseline. For the removal of QSA, we simply remove the branch and train an MLP to select the final mask prediction $\hat{M}$ from the proposed mask prediction $M_{m}$. For the removal of PWAM, we simply perform element-wise addition of the sentence features $F_{s}$ with the multi-scale features $\{F_{i}\}_{i=1}^{5}$. 

In this ablation study, we evaluate the effectiveness of the QSA and PWCA modules. As indicated in Table\ref{tab:module_abl}, the ablation model (a) only retain the sparse 3D feature extractor and query mask predictor. Here we perform element-wise addition of the word features $W$ to the multi-scale features $\{V_{i}\}_{i=1}^{5}$ instead of PWCA and project the proposed mask prediction $M_{m} \in \mathbb{R}^{K \times N}$ to the final mask prediction $\hat{M} \in \mathbb{R}^{N}$ via an MLP instead of QSA. We set model (a) as the baseline of our experiment. Compared to model (a), model (b) adopts PWCA module and we find that the model performance increase greatly from 32.66\% to 33.44\%. This observation proves that PWCA facilitate fine-grained cross-modal alignment between points and words, which is more effectively to leverage the rich convolutional layers in the encoder to excavate multi-modal context. When we introduce QSA to model (b), as shown in model (c), we can find that the performance of mIoU is improved from 33.44\% to 33.74\%. Such results indicates that the fine-grained point-word alignment of PWCA and the coarse-grained query-sentence alignment of QSA effectively coupled to enhance the capability in capturing multi-modal context.

\paragraph{Loss Ablation.}
As demonstrated in \ref{section:Loss}, we refine predicted masks from coarse to fine by introducing two loss functions, i.e., area regularization loss $\mathcal{L}_{area}$ and point-to-point contrastive loss $\mathcal{L}_{p2p}$. We successively add the loss functions and the results are shown in Tab.\ref{tab:loss_abl}. The model (a) is only supervised by segmentation loss $\mathcal{L}_{seg}$. When we introduce the $\mathcal{L}_{area}$ into model (a), as shown in model (b), the performance greatly increase from 25.86\% to 31.04\% on mIoU, which indicates our area loss can exclude a significant number of irrelevant background points. Moreover, we successively add the contrastive loss $\mathcal{L}_{p2p}$,  as shown in the model (c). The performance is improved from 31.04\% to 33.74\%, which proves that the contrastive loss can make the model more focused on the target area rather than others.

Qualitative results are shown in Fig.\ref{fig:ablation}, it indicates that: \textbf{i)} When only the segmentation loss $\mathcal{L}_{seg}$ is applied \textit{(column 3)}, the predicted masks include many points from other regions. \textbf{ii)} After adding the $\mathcal{L}_{area}$, most of the irrelevant points disappear \textit{(row a, row d)}. \textbf{iii)} After incorporating the $\mathcal{L}_{p2p}$, objects that were previously difficult to distinguish due to their similarity are successfully separated, and the predictions are close to the ground-truth.

Both quantitative and qualitative experiment demonstrate that our proposed loss function effectively reduces large-scale background misclassifications and distinguish objects or points with the similar characteristics.

\subsection{Extension Experiments}

\begin{table}[!t]
\caption{The impact of linguistic features at different granularities. \textbf{Word} represent word-level features, and \textbf{Sentence} represent sentence-level features.}
\label{tab:extension_language}
\centering
\begin{tabular}{cc|ccc}
\toprule
Word & Sentence & mIoU & Acc@0.25 & Acc@0.5 \\ \midrule
\checkmark  &  & 31.56&	49.50&	25.84\\
 & \checkmark & 33.02&	52.33 &	27.75  \\
\checkmark & \checkmark & \textbf{33.74} & \textbf{53.23} & \textbf{29.88} \\
\bottomrule
\end{tabular}
\end{table}

% \begin{table*}
% \begin{floatrow}
% \capbtabbox{
% \scalebox{0.9}{
% \begin{tabular}{cccc|ccc}
% \toprule
%  & Back & Mid & End & mIoU & A@25 & A@50  \\ \midrule
% (a) & &  & \checkmark & 14.84 & 18.83 &  4.58  \\
% (b) & & \checkmark &  & 27.00 & 43.68 & 19.30   \\
% (c) &\checkmark &  &  & 33.44 & 52.73 & 28.92    \\ \midrule
% (d) & \checkmark & \checkmark &   & 33.24 & 52.88 & 28.07   \\
% (e) &\checkmark &  & \checkmark &  \textbf{33.74} & \textbf{53.23} & \textbf{29.88}  \\
% (f) & \checkmark & \checkmark & \checkmark & 33.35 &  & \\
% \bottomrule
% \end{tabular}
% }}{
%  \caption{Multi-modal interactions at different position of the network. \textbf{Back} denotes interaction within the visual backbone, \textbf{Mid} represents interaction between the visual backbone and the QMP module, and \textbf{End} refers to interaction at QSA module.}
%  \label{tab:extension_position}
%  \small
% }
% \capbtabbox{
% \scalebox{0.9}{
% \begin{tabular}{cc|ccc}
% \toprule
% W. & S. & mIoU & A@25 & A@50 \\ \midrule
% \checkmark  &  & 31.56&	49.50&	25.84\\
%  & \checkmark & 33.02&	52.33 &	27.75  \\
% \checkmark & \checkmark & \textbf{33.74} & \textbf{53.23} & \textbf{29.88} \\
% \bottomrule
% \end{tabular}
% }}{
%  \caption{The impact of linguistic features at different granularities. \textbf{W.} represent word-level features, and \textbf{S.} represent sentence-level features.}
%  \label{tab:extension_language}
%  \small
% }
% \end{floatrow}
% \end{table*}

\begin{table*}
\begin{floatrow}
\capbtabbox{
\begin{tabular}{c|ccc}
\toprule
Num of queries  & mIoU & A@25 & A@50  \\ \midrule
20  & \textbf{33.74} & \textbf{53.23} & \textbf{29.88}  \\
60 & 32.62 & 51.96 & 27.70 \\
100  & 33.16 & 52.98 & 28.82  \\
\bottomrule
\end{tabular}
}{
 \caption{The impact of the number of queries.}
 \label{tab:sensitivity_query}
 \small
}
\capbtabbox{
\begin{tabular}{c|ccc}
\toprule
Num of layers  & mIoU & A@25 & A@50 \\ \midrule
1  & \textbf{33.74} & \textbf{53.23} & \textbf{29.88}  \\
3 & 32.45 & 51.09 & 27.70 \\
6  & 32.66 & 52.18 & 28.33  \\
\bottomrule
\end{tabular}
}{
 \caption{The impact  of the number of layers.}
 \label{tab:sensitivity_layer}
 \small
}
\end{floatrow}
\end{table*}

\paragraph{Linguistic Features at Different Levels of Granularity.}

% 虽然大部分的3DVG方法都同时使用了word-level的特征和sentence-level的特征，但目前的R33DIS方法只使用词level的特征进行多模态特征融合。因此，在本实验中，我们探究句level的特征是否适合用于Referring 3D Segmentation这个细粒度的任务。我们保持模型的结构不变，只改变语言特征的粒度，如表5所示，.....
% 这可能是由于句特征拥有更完整的上下文信息，并且不会对
In this section, we will investigate the impact of linguistic features at different levels of granularity, as shown in Table.\ref{tab:extension_language}. The first row represent that we leverage word-level features in both PWCA module and QSA module as text features. The second row indicates we utilize sentence-level features in both modules. We can find that sentence-only method even outperforms the word-only one. Referring 3D Segmentation task involves the segmentation of a solitary target object. This mandates a more profound and thorough comprehension of the semantic information conveyed by the sentence, extending beyond a mere focus on its individual words. The last row indicates the word-level features  are utilized in PWCA module and sentence-level features are leveraged in QSA module. We can find that the performance of it outperforms two introduced above, which proves that fine-grained word-level feature is also helpful in extracting the 3D language-aware visual feature.

% Even though current 3D visual grounding methods \cite{multiview3dvg, multi3drefer} utilize both word-level and sentence-level features, the exist Referring 3D Segmentation methods only use word-level features for multi-modal fusion. Therefore, in this experiment, we investigate whether sentence-level features are suitable for Referring 3D Segmentation such a fine-grained task. We keep the model architecture, altering only the granularity of the linguistic features. As shown in Tab.\ref{tab:extension_language}, the first row and second row represent that we use the word-level features and sentence-level features in both PWCA module and QSA module respectively. We can find that sentence-only methods even outperform the word-only one. Referring 3D Segmentation involves the segmentation of a solitary target object. This mandates a more profound and thorough comprehension of the semantic information conveyed by the sentence, extending beyond a mere focus on its individual phrases. The last row indicates the word-level features utilized in PWCA module and sentence-level features leveraged in QSA module and the performance of it outperforms two introduced above, which the proves the fine-grained word-level feature is also helpful in extracting the 3D  language-aware visual feature.

\paragraph{Layer and Query Number of QMP.}

We also investigate the performance on different query and layer numbers in the QMP module. As shown in Tab.\ref{tab:sensitivity_query} and Tab.\ref{tab:sensitivity_layer}, we find that too many queries and layers do not bring performance gains for Referring 3D Segmentation task. As a result, we set 20 queries and 1 QMP layer as the default configuration after balancing performance and efficiency.

\subsection{Limitations}

The limitations of LESS due to the inherent complexity of 3D point clouds and the ambiguous queries, although we have made significant improvements on previous methods. The scarcity of detailed semantic annotations are still challenging our model from distinguishing multiple similar objects. These limitations could guide our future work.

\section{Conclusions}

% 通过......，我们实现了......，详细的实验和消融结果表明.....，然而，我们的局限性在...

In this paper, we propose LESS, a label-efficient and single-stage approach for Referring 3D Segmentation. Specifically, our LESS enhances feature extraction by integrating multi-modal features and employs progressive constraints on predicted masks, achieving fine-grained alignment between points and words and distinguishing between points or objects with similar characteristics. Comprehensive experiments demonstrate that our single-stage method outperforms existing two-stage approaches on ScanRefer dataset, using only the binary labels as supervision. Though our framework still has some limitations, we believe that solving the Referring 3D Segmentation task only using the binary labels is a new and promising research, and we hope that LESS can serve as a simple but powerful baseline to inspire future research on Referring 3D Segmentation. 
% However, due to the inherent complexity of 3D point clouds and the scarcity of detailed semantic annotations, the ambiguous queries and scenarios with dense multiple objects may confuse our model.

\section{Acknowledgements}

This work was supported in part by the National Natural Science Foundation of China under Grants 62371310, in part by the Guangdong Basic and Applied Basic Research Foundation under Grant 2023A1515011236, in part by the Stable Support Project of Shenzhen (Project No.20231122122722001).

We also thank the support by the MUR PNRR project FAIR (PE00000013) funded by the NextGenerationEU, the PRIN project CREATIVE (Prot. 2020ZSL9F9), and the EU Horizon project ELIAS (No. 101120237). We also acknowledge the CINECA award under the ISCRA initiative, for the availability of partial HPC resources support.

\bibliographystyle{plain}
\bibliography{references}

%%%%%%%%%%%%%%%%%%%%%%%%%%%%%%%%%%%%%%%%%%%%%%%%%%%%%%%%%%%%
\newpage
\appendix

\section*{Appendix}

\section{Semantic and Instance Labels vs. Binary Labels}

\begin{figure}[h]
  \centering
  \includegraphics[width=\linewidth]{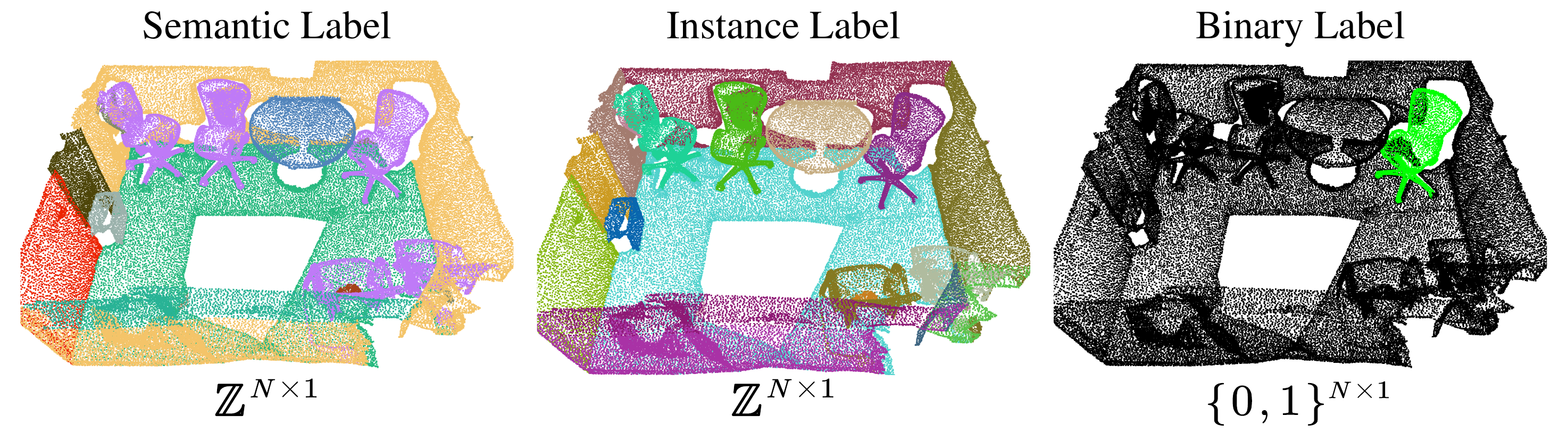}
  \caption{There different types of labels. }
  \label{fig:label}
\end{figure}

The differences between semantic labels, instance labels, and binary labels are shown in Fig.\ref{fig:label}.

Semantic labels assign a category ID to each point in a scene, with points belonging to the same category (e.g., chairs) sharing the same ID. Instance labels assign a object ID to each point within the same object (e.g., a specific chair), distinguishing different objects by assigning different instance IDs. Binary labels assign a Boolean value to each point in the scene; points that are part of the described object are assigned a value of 1, and points that are not are assigned a value of 0.

Therefore, compared to binary labels, the annotations of semantic labels and instance labels are more time consuming and labor intensive.

\section{More Qualitative Results}

\begin{figure}[h]
  \centering
  \includegraphics[width=\linewidth]{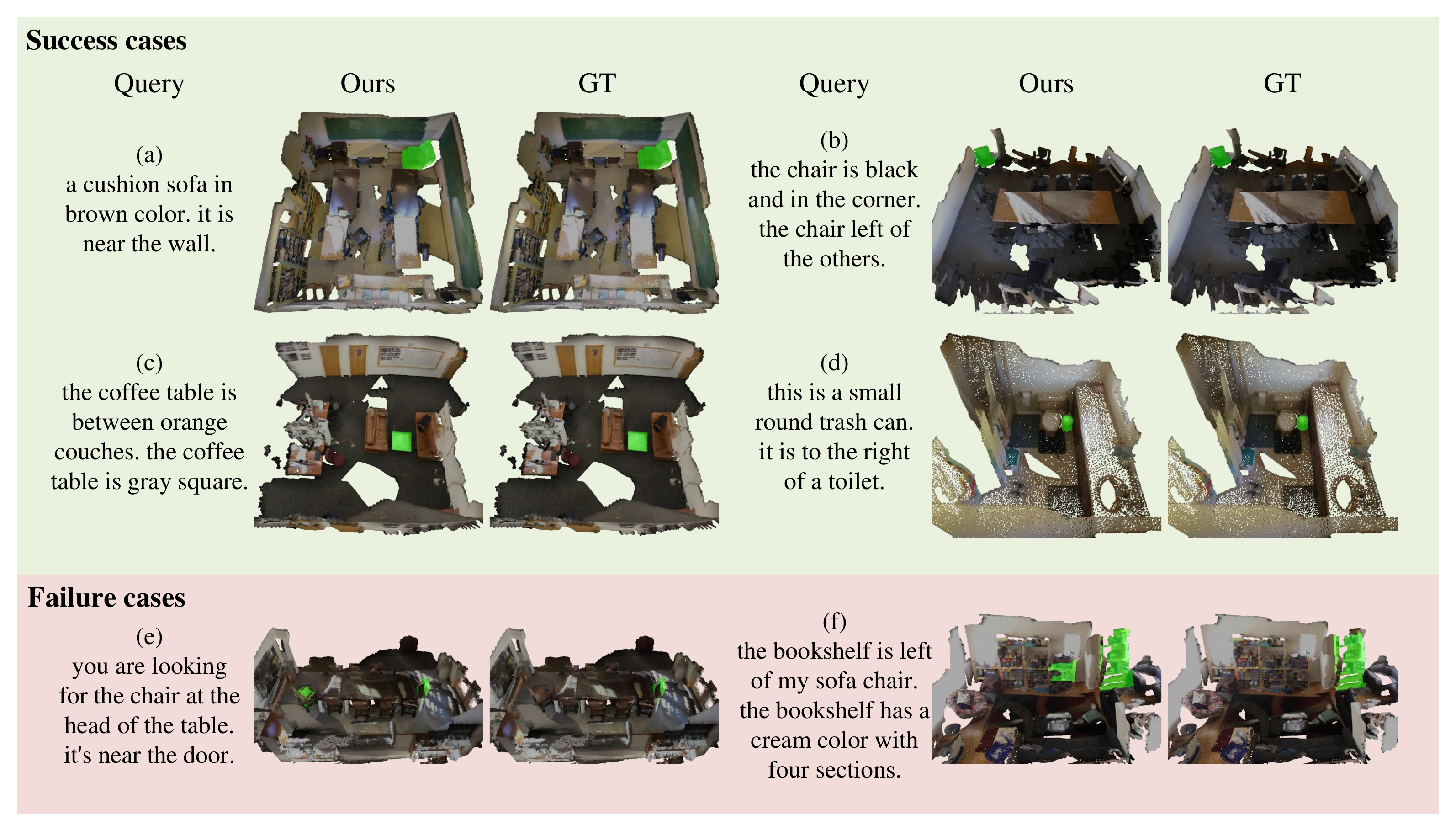}
  \caption{Qualitative results of both the success cases and failure cases of our LESS.}
  \label{fig:visualization}
\end{figure}

We present our success cases and failure cases in Fig.\ref{fig:visualization}. Our method accurately segments objects with clear queries. However, ambiguous descriptions can still confuse our model and leads to segment both the referred object and other similar objects.

\section{Time Comsumption Comparison}

%单阶段的优势在于我们可以同时进行点云的特征提取和多模态的信息融合%

As shown in Tab.\ref{tab:time}, we evaluate the training and inference time both of TGNN\cite{TGNN} and X-RefSeg\cite{Xrefer}. All experiments are conducted on an NVIDIA 4090 GPU and the number of batch sizes and epoch of three methods are kept the same. For the two-stage training and inference of TGNN and X-RefSeg, we followed the settings in their open source codes. We can find that our LESS consumes less time than both of TGNN and X-RefSeg in both training and inference.

\begin{table}[h]
\caption{The comparison of inference time and training time with previous work.}
\label{tab:time}
\centering
\begin{tabular}{cccccc}
\toprule
Method   & \begin{tabular}[c]{@{}c@{}}Inference \\ (Whole Dataset) (min)\end{tabular} & \begin{tabular}[c]{@{}c@{}}Inference \\ (Per Scan) (ms)\end{tabular} & \begin{tabular}[c]{@{}c@{}}Training \\ (Stage 1) (h)\end{tabular} & \begin{tabular}[c]{@{}c@{}}Training \\ (Stage 2) (h)\end{tabular} & \begin{tabular}[c]{@{}c@{}}Training \\ (All) (h)\end{tabular} \\ \midrule
TGNN     & 27.98                                                                     & 176.57                                                              & 156.02                                                           & 8.53                                                             & 164.55                                                       \\
X-RefSeg & 20.00                                                                     & 126.23                                                              & 156.02                                                           & 7.59                                                             & 163.61                                                       \\
Ours     & \textbf{7.09}                                                             & \textbf{44.76}                                                      & -                                                                & -                                                                & \textbf{40.89}                                               \\
                                            
\bottomrule
\end{tabular}
\end{table}

\section{More Quantitative Results}

A concurrent work 3D-STMN \cite{3dstmn} utilizes a pre-trained 3D feature extractor \cite{spformer} to perform referring 3D segmentation. Given that their backbone is pre-trained on an instance segmentation task with semantic and instance label, it is reasonable to conclude that their approach cannot be considered a label-efficient and single-stage method. For fair comparison, we follow the settings in their open source code and train their network from scratch, except for BERT \cite{bert} module. As shown in Tab.\ref{tab:3dstmn}, our method overtakes 3D-STMN by 11.34\%, 18.27\% and 12.65\% on mIoU, Acc@0.25 and Acc@0.5 respectively.

\begin{table}[h]
\caption{Quantitative results of 3D-STMN \cite{3dstmn} and ours.}
\label{tab:3dstmn}
\centering
\begin{tabular}{cccc}
\toprule
Method                & mIoU  & Acc@0.25 & Acc@0.5 \\ \midrule
3D-STMN (from scratch) & 22.40 & 34.96  & 17.23  \\
Ours     & \textbf{33.74} & \textbf{53.23} & \textbf{29.88}  \\ \bottomrule
\end{tabular}
\end{table}

\section{More Ablation Results}

\subsection{Query Number}
We further conduct more ablation studies in terms of the number of queries on ScanRefer. As shown in Tab.\ref{tab:sensitivity_query2}, we ablate the number of queries as {5, 15, 20}, it can be observed that keeping 20 queries brings higher accuracy while lower accuracy when using fewer queries. We suppose that fewer queries can not help to learn comprehensive feature patterns while an appropriate number of queries is enough to cover the needed feature patterns.

\begin{table}[h]
\caption{The impact of the number of queries.}
\label{tab:sensitivity_query2}
\begin{tabular}{c|ccc}
\toprule
Num of queries  & mIoU & Acc@0.25 & Acc@0.5  \\ \midrule
5 & 32.75 & 51.81 & 28.68 \\
15 & 33.27 & 52.56 & 28.97 \\
20  & \textbf{33.74} & \textbf{53.23} & \textbf{29.88}  \\
\bottomrule
\end{tabular}
\end{table}

\subsection{Mask Selection Strategy}

We also conduct the ablation study which selects the mask with the highest score during QSA process as shown in Tab.\ref{tab:maskselection}. The results demonstrate that the aggregation of multiple masks yields superior performance compared to the selection of a single, highest-ranked mask. This approach facilitates the capture of subtle nuances and intricate details that may be overlooked when relying on a single mask alone. The incorporation of multiple masks offers a more comprehensive and precise representation, ultimately enhancing the accuracy of the final model prediction.

\begin{table}[h]
\caption{The impact of mask selection strategy.}
\label{tab:maskselection}
\begin{tabular}{c|ccc}
\toprule
Num of queries  & mIoU & Acc@0.25 & Acc@0.5  \\ \midrule
Top-1 & 33.18 & 52.93 & 28.65 \\
Ours  & \textbf{33.74} & \textbf{53.23} & \textbf{29.88}  \\
\bottomrule
\end{tabular}
\end{table}

\end{document}